\documentclass{article} 
\usepackage{iclr2026_conference,times}

\usepackage{amsmath,amsfonts,bm}

\def\eqref#1{equation~\ref{#1}}

\def\1{\bm{1}}

\DeclareMathAlphabet{\mathsfit}{\encodingdefault}{\sfdefault}{m}{sl}
\SetMathAlphabet{\mathsfit}{bold}{\encodingdefault}{\sfdefault}{bx}{n}

\usepackage{booktabs}
\usepackage{graphicx}
\usepackage{placeins}
\usepackage{hyperref}
\usepackage{url}

\newcommand{\method}{QK-Normed MLA}

\title{\method: QK Normalization Without Full Key Caching}

\author{
\vspace{0.5em}\\
\makebox[\dimexpr\textwidth-2\tabcolsep\relax][c]{%
\begin{tabular}{@{}c@{\hspace{2.1em}}c@{\hspace{2.1em}}c@{}}
Yizhou Han\textsuperscript{1,2} & Yao Zhao\textsuperscript{2} & Jun Zhou\textsuperscript{2} \\
{\small\texttt{223040226@link.cuhk.edu.cn}} &
{\small\texttt{nanxiao.zy@antgroup.com}} &
{\small\texttt{jun.zhoujun@antgroup.com}} \\[8pt]
\multicolumn{3}{c}{%
\begin{tabular}{@{}c@{\hspace{3em}}c@{}}
Longfei Li\textsuperscript{2}\thanks{Corresponding authors.} &
Ruoyu Sun\textsuperscript{1}\textsuperscript{*} \\
{\small\texttt{longyao.llf@antgroup.com}} & {\small\texttt{sunruoyu@cuhk.edu.cn}}
\end{tabular}%
}
\end{tabular}%
}
}

\iclrfinalcopy
\begin{document}

\maketitle
\footnotetext[1]{School of Data Science, The Chinese University of Hong Kong, Shenzhen.}
\footnotetext[2]{Ling Team, AI@Ant Group.}
\lhead{}
\chead{}
\rhead{}

\begin{abstract}
Query-key (QK) normalization stabilizes attention by controlling the scale of
queries and keys before the dot product, but is not immediately compatible with
Multi-head Latent Attention (MLA). MLA achieves efficient decoding by caching
low-dimensional latent states instead of full keys, whereas post-projection QK
RMSNorm appears to require the fully projected key for every cached token. We show
this apparent incompatibility is an implementation artifact, not an architectural
constraint.
RMSNorm decomposes into a static affine weight and a dynamic scalar RMS statistic.
The static key-side weight can be absorbed into the MLA query-side projection; the
dynamic key statistic reduces to one inverse-RMS scalar per token and KV group. The
resulting formulation is exactly equivalent to explicit post-projection QK RMSNorm in
exact arithmetic and preserves MLA's latent decode path. In our 400M runs trained for
up to 100B tokens, \method{} achieves lower training loss and better downstream
accuracy than QK clipping, while H800 decode benchmarks show less than $2\%$ latency
overhead up to 256k context. These results make QK normalization a
practical stabilization option for MLA models without requiring full-key caching.
\end{abstract}

\section{Introduction}

Two practical pressures shape modern language-model attention. The first is KV-cache
efficiency. Standard multi-head attention~\citep{vaswani2017attention} stores a key
and value vector for every previous token in every layer, making KV-cache memory and
bandwidth central costs in long-context inference. Multi-query attention and
grouped-query attention reduce the number of KV heads
~\citep{shazeer2019onewritehead,ainslie2023gqa}. Multi-head Latent Attention (MLA)
goes further: it caches a compact latent state and recovers the effect of full
content keys by moving the key up-projection to the query side during decoding
~\citep{deepseekv2,deepseekv3}.

The second pressure is training stability. QK normalization controls attention-logit
scale by normalizing queries and keys along the per-head feature dimension before
forming their dot product~\citep{henry2020qknorm}. This is attractive because
uncontrolled query/key scale can lead to saturated attention, loss spikes, or hard
training instabilities. A different practical response is to clip large QK logits, as in
MuonClip/QK-Clip~\citep{kimik2}. Clipping acts after the dot product, while QK
normalization removes a scale degree of freedom before the attention geometry is
formed.

This creates a practical compatibility gap between two increasingly important
ingredients. QK normalization is a direct stabilizer for large-scale training, while
MLA is a memory-saving attention design for efficient decoding and long-context use.
The missing piece is that post-projection QK RMSNorm appears to require exactly the
full keys that MLA is designed not to cache. Concretely, the efficient MLA decode
path avoids caching full content keys, but post-up-projection QK RMSNorm seems to
require normalizing the key
\[
    k^0_{t,g} = c_t W^K_g
\]
for every cached token $t$ and KV group $g$. A direct implementation would either
materialize normalized full keys or expand the cache in a way that gives up much of
MLA's memory advantage. The question is whether MLA can use standard post-projection
QK RMSNorm without abandoning latent decoding.

We show that this is possible. The key observation is that RMSNorm separates into a
static feature-wise affine weight and a dynamic scalar RMS statistic:
\[
    \operatorname{RMSNorm}(z)
    =
    \frac{z \odot \gamma}{S(z)},
    \qquad
    S(z)
    =
        \sqrt{\frac{1}{d}\sum_{j=1}^{d} z_j^2 + \epsilon}.
\]
The static key-side weight $\gamma$ can be absorbed into the query-side MLA
projection. The dynamic scale $S(z)$ cannot be absorbed into a fixed matrix, but for
keys it is only one scalar per token and KV group. Therefore the decode cache can keep
the original latent state $c_t$ and add a small inverse-RMS scalar cache. During
attention, the latent dot product is multiplied by the corresponding query and key
RMS factors before softmax. This gives the same logits as explicitly materializing
the post-up-projection key and applying QK RMSNorm, up to ordinary floating-point
roundoff.

This paper makes three contributions.

\begin{itemize}
\item We derive an exact formulation of post-up-projection QK RMSNorm that is
compatible with MLA's latent decode path.
\item We give a cache layout and decode data flow that preserve the latent KV cache
as the dominant state, adding only one scalar key RMS statistic per token and KV
group.
\item We observe lower training loss and better downstream accuracy for \method{}
over QK clipping in 400M models trained for 100B tokens, while adding less than
$2\%$ decode latency in our H800 benchmark.
\end{itemize}

The main claim is algebraic and implementation-oriented: standard post-projection QK
RMSNorm is compatible with efficient MLA decoding. We use max-logit control as a
mechanism-level diagnostic, but evaluate the method primarily through training loss,
downstream accuracy, and decode-time overhead.

\section{Background}

\subsection{Query-Key Normalization}

For a query head $h$ attending to a key group $g$, standard scaled dot-product
attention forms logits
\[
    \ell_{i,t,h}
    =
    \frac{q_{i,h}\cdot k_{t,g(h)}}{\sqrt{d_h}},
\]
where $i$ is the query position, $t$ is a key position, and $g(h)$ maps a query head
to its KV head or group. QK normalization instead normalizes $q$ and $k$ along the
per-head feature dimension before the dot product~\citep{henry2020qknorm}. In the
RMSNorm form~\citep{zhang2019rmsnorm}, this is
\begin{equation}
    \hat q_{i,h}
    =
    \frac{q^0_{i,h}\odot\gamma_q}{S^q_{i,h}},
    \qquad
    \hat k_{t,g}
    =
    \frac{k^0_{t,g}\odot\gamma_k}{S^k_{t,g}},
    \label{eq:qk-rmsnorm}
\end{equation}
with
\[
    S^q_{i,h}
    =
    \sqrt{\frac{1}{d_h}\sum_{j=1}^{d_h}(q^0_{i,h,j})^2+\epsilon},
    \qquad
    S^k_{t,g}
    =
    \sqrt{\frac{1}{d_h}\sum_{j=1}^{d_h}(k^0_{t,g,j})^2+\epsilon}.
\]
In common implementations, the RMS statistic is per token and per head, while the
affine weight $\gamma_q,\gamma_k\in\mathbb{R}^{d_h}$ is shared across heads and
broadcast over the head axis.

\subsection{Multi-head Latent Attention}

MLA introduces a low-dimensional latent cache. For each token $t$, an input hidden
state $x_t$ is projected to a latent vector
\begin{equation}
    c_t = x_t W_D,
    \qquad
    c_t\in\mathbb{R}^{r},
    \label{eq:latent}
\end{equation}
where $r$ is smaller than the total full-key dimension. A content key for KV group
$g$ is obtained by an up projection
\begin{equation}
    k^{0,c}_{t,g}
    =
    c_t W^K_g,
    \qquad
    W^K_g\in\mathbb{R}^{r\times d_c}.
    \label{eq:key-up}
\end{equation}
Ignoring RoPE for the moment, a content attention score is
\[
    \ell^c_{i,t,h}
    =
    q^{0,c}_{i,h}\cdot k^{0,c}_{t,g(h)}
    =
    q^{0,c}_{i,h}\cdot(c_t W^K_{g(h)}).
\]
By associativity,
\begin{equation}
    q^{0,c}_{i,h}\cdot(c_t W^K_{g(h)})
    =
    \left(q^{0,c}_{i,h}(W^K_{g(h)})^\top\right)\cdot c_t.
    \label{eq:mla-absorb}
\end{equation}
Thus decoding can transform the current query into latent space and dot it with the
cached $c_t$, without storing the materialized full key $k^{0,c}_{t,g}$ for every
past token.

\section{Method}

\subsection{Key-Side RMSNorm Without Caching Full Keys}

Consider post-up-projection key RMSNorm on the MLA content key:
\begin{equation}
    \hat k^c_{t,g}
    =
    \frac{
        (c_t W^K_g)\odot\gamma^c_k
    }{
        S^k_{t,g}
    },
    \label{eq:key-rms}
\end{equation}
where
\[
    S^k_{t,g}
    =
    \sqrt{
        \frac{1}{d_c}
        \sum_{j=1}^{d_c}
        (c_t W^K_g)_j^2
        + \epsilon
    }.
\]
The normalized-key content score is
\[
    \ell^c_{i,t,h}
    =
    q^{0,c}_{i,h}\cdot \hat k^c_{t,g(h)}.
\]
Substituting \eqref{eq:key-rms},
\begin{align}
    \ell^c_{i,t,h}
    &=
    q^{0,c}_{i,h}
    \cdot
    \frac{
        (c_t W^K_{g(h)})\odot\gamma^c_k
    }{
        S^k_{t,g(h)}
    }
    \nonumber \\
    &=
    \left(
        (q^{0,c}_{i,h}\odot\gamma^c_k)
        (W^K_{g(h)})^\top
    \right)
    \cdot c_t
    \cdot
    \frac{1}{S^k_{t,g(h)}}.
    \label{eq:key-absorbed}
\end{align}
Define
\begin{equation}
    \tilde q^c_{i,h}
    =
    (q^{0,c}_{i,h}\odot\gamma^c_k)
    (W^K_{g(h)})^\top,
    \qquad
    \alpha^k_{t,g}
    =
    \frac{1}{S^k_{t,g}}.
    \label{eq:key-defs}
\end{equation}
Then
\begin{equation}
    \ell^c_{i,t,h}
    =
    (\tilde q^c_{i,h}\cdot c_t)
    \alpha^k_{t,g(h)}.
    \label{eq:key-final}
\end{equation}
Equation~\ref{eq:key-final} is the key observation. The affine RMSNorm weight
$\gamma^c_k$ is static and enters the query-side projection. The only key-dependent
runtime quantity that remains is the scalar $\alpha^k_{t,g}$.

Therefore the decode cache does not need normalized full keys. It stores
\[
    C_{\mathrm{cache}}
    =
    \{c_t\}_{t=1}^{T},
    \qquad
    A^k_{\mathrm{cache}}
    =
    \{\alpha^k_{t,g}\}_{t=1,g=1}^{T,G},
\]
where $G$ is the number of KV groups. The additional state is $TG$ scalars per layer,
small compared with caching $TGd_c$ full content keys.

\subsection{Query-Side RMSNorm}

Query RMSNorm can be handled in the same factorized logit form. For the content query,
\begin{equation}
    \hat q^c_{i,h}
    =
    \frac{q^{0,c}_{i,h}\odot\gamma^c_q}{S^q_{i,h}},
    \qquad
    \alpha^q_{i,h}
    =
    \frac{1}{S^q_{i,h}}.
    \label{eq:query-rms}
\end{equation}
Combining query and key RMSNorm gives
\begin{align}
    \ell^c_{i,t,h}
    &=
    \hat q^c_{i,h}\cdot \hat k^c_{t,g(h)}
    \nonumber \\
    &=
    \left(
        (q^{0,c}_{i,h}\odot\gamma^c_q\odot\gamma^c_k)
        (W^K_{g(h)})^\top
        \cdot c_t
    \right)
    \alpha^q_{i,h}
    \alpha^k_{t,g(h)}.
    \label{eq:qk-final}
\end{align}
Equivalently, an implementation may first compute the normalized query
$\hat q^c_{i,h}$ and then form
\[
    \tilde q^c_{i,h}
    =
    (\hat q^c_{i,h}\odot\gamma^c_k)
    (W^K_{g(h)})^\top.
\]
This avoids computing the query projection twice. The scalar $\alpha^q_{i,h}$ is
known for the current token and can be applied to all logits for head $h$.

\subsection{Content and RoPE Components}

In MLA-style architectures, the per-head query and key dimensions are often split
into a content component and a RoPE component:
\[
    q^0_{i,h}
    =
    [q^{0,c}_{i,h}; q^{0,r}_{i,h}],
    \qquad
    k^0_{t,g}
    =
    [k^{0,c}_{t,g}; k^{0,r}_{t}],
\]
with dimensions $d_c$ and $d_r$. The RoPE key component is materialized because it is
small and shared across heads or KV groups. We therefore use a blockwise
normalization scheme that follows the natural systems boundary of MLA.

Normalize the content and RoPE blocks separately:
\[
    \hat q^c
    =
    \operatorname{RMSNorm}^c_q(q^{0,c}),
    \qquad
    \hat k^c
    =
    \operatorname{RMSNorm}^c_k(k^{0,c}),
\]
\[
    \hat q^r
    =
    \operatorname{RMSNorm}^r_q(q^{0,r}),
    \qquad
    \hat k^r
    =
    \operatorname{RMSNorm}^r_k(k^{0,r}).
\]
The content path uses the absorbed latent formulation derived above. The RoPE path
uses ordinary materialized normalized vectors. The resulting score is
\begin{equation}
    \ell_{i,t,h}
    =
    \left(\tilde q^c_{i,h}\cdot c_t\right)
    \alpha^{q,c}_{i,h}
    \alpha^{k,c}_{t,g(h)}
    +
    \hat q^r_{i,h}\cdot \hat k^r_t .
    \label{eq:blockwise-final}
\end{equation}
This is not the only algebraically possible normalization. One could normalize the
concatenated content--RoPE vector with a unified RMS statistic. We do not use that
variant because it is a poor fit for efficient training implementations: unified
normalization introduces a synchronization point between the content and RoPE compute
paths, complicating the fused kernel structure that makes MLA efficient.
Appendix~\ref{app:blockwise-vs-unified} gives the detailed systems rationale. In the
main method, the important point is that blockwise QK RMSNorm preserves the
implementation boundary that makes MLA efficient: the content component remains latent,
while the RoPE component remains small and materialized.

\subsection{Decode-Time Data Flow}

The absorbed formulation changes where normalization statistics are stored and where
the scalar factors are applied, but it does not change the dominant cache object.
For each new token $t$, \method{} writes the cache as follows.
\begin{enumerate}
\item Compute the latent state $c_t=x_tW_D$ and store it in $C_{\mathrm{cache}}$.
\item Compute the temporary content-key matrix
$K^{0,c}_t=c_tW^K\in\mathbb{R}^{G\times d_c}$ in the same dense projection used by
the content path, reduce each group to obtain $S^{k,c}_{t,g}$, and discard
$K^{0,c}_t$.
\item Store $\alpha^{k,c}_{t,g}=1/S^{k,c}_{t,g}$ in the scalar cache.
\item Normalize the small materialized RoPE key component inside the RoPE path and
store it as in standard MLA.
\end{enumerate}
The temporary content-key matrix in the second step is used only to compute RMS
statistics; it is not written as a historical key cache. In common layouts, the key
up-projection is a dense matrix
\[
    W^K\in\mathbb{R}^{r\times Gd_c},
\]
whose output is reshaped to $G\times d_c$ only for the reduction that produces the
RMS scalars. The cache writer therefore keeps the large operation in the dense GEMM
path and stores only the latent vector plus $G$ scalar inverse RMS values.

For a later query position $i$, compute the content-query RMS statistics
$\alpha^{q,c}_{i,h}$ and form the latent-space query with the static RMSNorm weights
absorbed:
\[
    \tilde q^c_{i,h}
    =
    \left(
        q^{0,c}_{i,h}
        \odot \gamma^c_q
        \odot \gamma^c_k
    \right)
    (W^K_{g(h)})^\top .
\]
The raw content score tile is then
\[
    L^{c,\mathrm{raw}}_{h,t}
    =
    \tilde q^c_{i,h}\cdot c_t,
\]
and the normalized content score is obtained by row--column scaling:
\begin{equation}
    L^c_{h,t}
    =
    L^{c,\mathrm{raw}}_{h,t}
    \alpha^{q,c}_{i,h}
    \alpha^{k,c}_{t,g(h)}.
    \label{eq:score-stage-scaling}
\end{equation}
This row--column scaling is best fused into the attention-score kernel, before the
online softmax max/sum update. The kernel loads a tile of $C_{\mathrm{cache}}$, the matching
tile of $A^k_{\mathrm{cache}}$, and the current query scalar $\alpha^{q,c}_{i,h}$;
after the latent dot-product accumulator is formed, it applies these factors and then
continues with the ordinary attention computation.

The scalar cache should remain separate from the latent cache. Since
$S^{k,c}_{t,g}$ differs across KV groups, absorbing it into the cached latent vector
would expand the cache from $T\times r$ to $T\times G\times r$, eliminating much of
MLA's memory advantage. The scalar formulation keeps the large cache latent and pays
only a small score-stage scaling cost.

\section{Analysis}

\subsection{Cache and Compute Cost}

The transformation preserves the dominant memory structure of MLA. Relative to
standard absorbed MLA decoding, \method{} adds only the scalar cache
\[
    A^k_{\mathrm{cache}}\in\mathbb{R}^{T\times G}.
\]
The dominant content cache remains the latent cache
\[
    C_{\mathrm{cache}}\in\mathbb{R}^{T\times r}.
\]
Since $r\gg G$ in typical MLA configurations, the extra state is small. For example,
when $r=512$ and $G=8$, the key-RMS cache is $1.56\%$ of the latent content cache
before accounting for values or RoPE state.

The extra compute is also localized. Relative to a purely absorbed MLA decode path,
the cache writer must form the temporary current-token projection
$K^{0,c}_t=c_tW^K$ sufficiently to compute the groupwise RMS statistics. This
projection is independent of the historical sequence length $T$. Attention over the
prefix still uses latent dot products with $C_{\mathrm{cache}}$; the only additional
prefix-dependent operation is scalar multiplication of the score accumulator.

\subsection{Why This Controls Logit Scale}

We do not claim that QK RMSNorm by itself defines an optimization theory, but it gives
a direct mechanism for controlling a common failure mode. Without normalization, a content
logit can grow with the product of the raw query and key norms:
\[
    |\ell^c_{i,t,h}|
    \le
    \lVert q^{0,c}_{i,h}\rVert_2
    \lVert k^{0,c}_{t,g(h)}\rVert_2.
\]
RMSNorm removes this direct dependence on the raw vector RMS. After normalization,
the effective norms are controlled primarily by the learned affine weights:
\[
    \lVert \hat q^c_{i,h}\rVert_2
    \approx
    \lVert \gamma^c_q\rVert_2,
    \qquad
    \lVert \hat k^c_{t,g}\rVert_2
    \approx
    \lVert \gamma^c_k\rVert_2,
\]
up to the directions of the unnormalized vectors and the $\epsilon$ term. This is why
we treat max-logit statistics as mechanism-level diagnostics: they indicate whether
the normalization is doing the intended scale control, while validation loss and
downstream evaluation remain the primary outcomes.

\subsection{Numerical Equivalence}

In exact arithmetic, the absorbed formulation is identical to explicitly materializing
the post-up-projection key and applying QK RMSNorm before the dot product. In
floating-point arithmetic, the two paths may differ slightly because they use
different multiplication and reduction orders, as fused and unfused attention kernels
already do. We therefore test implementation correctness by comparing explicit and
absorbed logits on random tensors before running training experiments; the reference
test is described in Appendix~\ref{app:equivalence-test}.

\section{Experiments}

The experiments evaluate three questions. First, can QK RMSNorm improve MLA training
relative to QK clipping under a controlled training setup? Second, does the method
produce the intended attention-scale behavior? Third, does the scalar cache preserve
MLA's decode efficiency in long-context settings? Training experiments use
400M-parameter MLA models trained for 100B tokens.

\subsection{Training Loss}

We compare QK-Clip and \method{} using the same architecture, training budget, data
mixture, optimizer schedule, and attention backend. The intended difference is the
attention stabilization mechanism: QK-Clip rescales Q/K weights when the observed
maximum logit exceeds a fixed threshold, whereas \method{} applies blockwise
post-up-projection RMSNorm to the content and RoPE QK components. A concise
configuration summary is provided in Appendix~\ref{app:experimental-config}.

\begin{figure}[h]
\begin{center}
\includegraphics[width=0.65\linewidth]{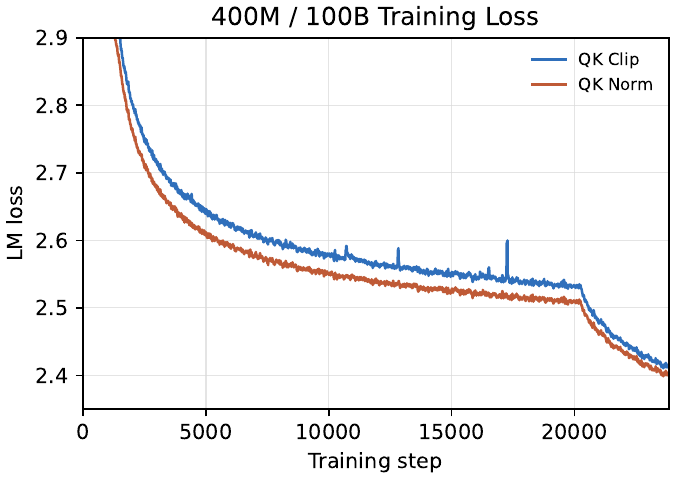}
\end{center}
\caption{Training loss of 400M MLA models trained for 100B tokens. \method{}
maintains consistently lower loss than QK-Clip throughout training.}
\label{fig:loss}
\end{figure}

Figure~\ref{fig:loss} shows the training loss over the full 100B-token run.
\method{} reaches lower loss throughout, and the gap persists over the entire
training horizon. This supports the view that QK normalization provides a persistent
training benefit over post-hoc clipping in this MLA setting. Diagnostic curves for maximum
attention logit and gradient norm (Appendix~\ref{app:diagnostics-100b}) are consistent
with the mechanism in Section~4.2: \method{} keeps logits in a lower range without
post-hoc intervention.

\subsection{Downstream Evaluation}

We evaluate 3-shot downstream performance with a standard language-model evaluation
suite. Table~\ref{tab:downstream} reports accuracy for each task, except LAMBADA
where we report accuracy and separately report perplexity. \method{} improves seven
of eight reported tasks, raises the average score from $44.75$ to $46.33$, and reduces
LAMBADA perplexity from $16.28$ to $14.18$.

\begin{table}[h]
\caption{Downstream evaluation of 400M/100B checkpoints. Perplexity (lower is better)
is reported separately; the average is computed over accuracy tasks only.}
\label{tab:downstream}
\begin{center}
\footnotesize
\setlength{\tabcolsep}{4.0pt}
\begin{tabular}{lc@{\hspace{8pt}}cccccccc@{\hspace{8pt}}c}
\toprule
Variant & \begin{tabular}[c]{@{}c@{}}Lamb.\\ppl$\downarrow$\end{tabular} &
ARC-C & ARC-E & BoolQ & HSwag & Lamb. & OBQA & PIQA & Wino & Avg.$\uparrow$ \\
\midrule
QK-Clip
& 16.28 & 22.61 & 53.07 & 59.51 & 38.32 & 42.40 & 18.20 & 70.67 & 53.20 & 44.75 \\
\method{}
& \textbf{14.18} & \textbf{24.66} & \textbf{56.69} & \textbf{62.20} & \textbf{38.94} & \textbf{43.92}
& \textbf{19.60} & \textbf{70.89} & \textbf{53.75} & \textbf{46.33} \\
\bottomrule
\end{tabular}
\end{center}
\end{table}

\subsection{Decode-Time Overhead}

We benchmark a single DeepSeek-V3-width MLA layer on H800 GPUs using BF16 and
\texttt{torch.compile(mode="reduce-overhead")}. The simulated TP=8 local view has
16 query heads, $r=512$ latent KV rank, $d_c=128$ content dimension, and $d_r=64$
RoPE dimension. The QK-normalized path stores the same latent KV cache and RoPE cache
as standard MLA, plus a local scalar RMS cache. Per layer and per GPU, this adds 2MB
to a 72MB MLA cache, a $2.8\%$ cache-size increase.

\begin{table}[h]
\caption{Decode-time overhead of adding QK RMS statistics to MLA. Times are
milliseconds per layer for batch size 4.}
\label{tab:inference}
\begin{center}
\begin{tabular}{lrrr}
\toprule
Context & MLA & +\method{} & Overhead \\
\midrule
4k   & 1.715 & 1.740 & 1.41\% \\
8k   & 1.891 & 1.909 & 0.95\% \\
16k  & 2.082 & 2.121 & 1.87\% \\
32k  & 2.486 & 2.511 & 1.03\% \\
64k  & 3.217 & 3.253 & 1.12\% \\
128k & 5.427 & 5.517 & 1.66\% \\
192k & 7.271 & 7.392 & 1.66\% \\
256k & 13.082 & 13.274 & 1.46\% \\
\bottomrule
\end{tabular}
\end{center}
\end{table}

Across context lengths from 4k to 256k, the measured decode overhead ranges from
$0.95\%$ to $1.87\%$, with an average overhead of $1.40\%$. This supports the
systems claim that, in this benchmark, the scalar cache and score-stage scaling do
not materially change MLA decode efficiency.

\FloatBarrier
\subsection{High-Learning-Rate Stress Test}

\begin{figure}[h]
\begin{center}
\includegraphics[width=0.92\linewidth]{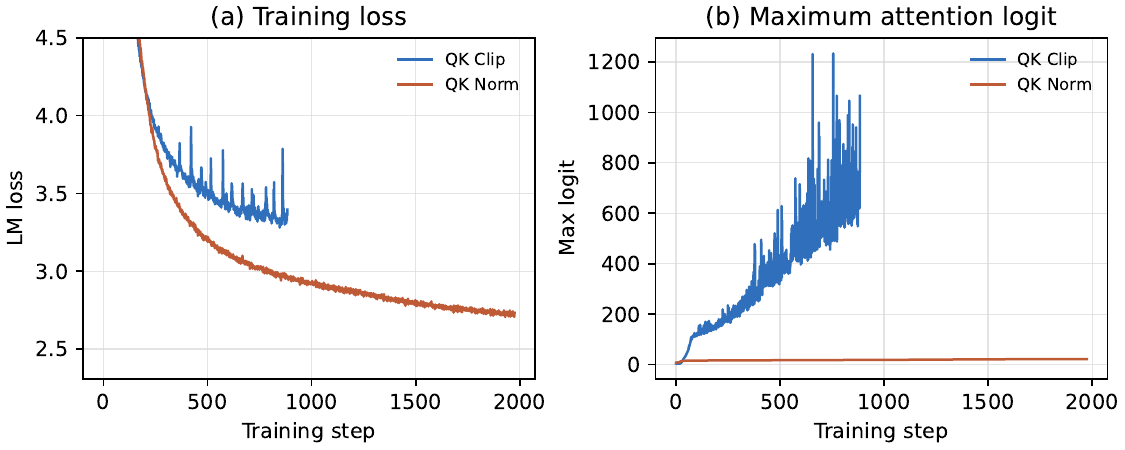}
\end{center}
\caption{High-learning-rate stress test ($\mathrm{lr}=2\times10^{-2}$). QK-Clip
diverges at step 884; \method{} remains stable over the same run.}
\label{fig:stress-test}
\end{figure}

As an additional instability test, we train an MLA model with an intentionally large
learning rate of $2\times 10^{-2}$. Under this stress setting, QK-Clip diverges with
gradient NaNs at step 884, as shown in Figure~\ref{fig:stress-test}. Before
divergence, its maximum attention logit grows rapidly and its loss fails to decrease
beyond the initial transient. \method{} remains stable over the same run and continues
to reduce loss. We use this only as a stress test; it is not intended as a recommended
training recipe.

\section{Related Work}

\paragraph{KV-efficient attention and serving.}
Transformer decoding is often limited by KV-cache bandwidth and capacity. Multi-query
attention shares one KV head across query heads~\citep{shazeer2019onewritehead}, and
grouped-query attention interpolates between multi-head and multi-query attention
~\citep{ainslie2023gqa}. Orthogonal systems work improves attention execution and
cache management through IO-aware kernels and paged KV allocation
~\citep{dao2022flashattention,kwon2023vllm}. MLA reduces the cache itself by storing
a low-dimensional latent state and moving the key up-projection to the query side
during decoding~\citep{deepseekv2,deepseekv3}. Our work keeps this latent-cache
design and asks how to combine it with post-projection QK normalization.

\paragraph{QK normalization and attention-logit control.}
QK normalization was proposed to reduce softmax saturation by normalizing queries and
keys before attention~\citep{henry2020qknorm}. Related large-model recipes use QK
normalization or logit control to improve training stability, including ViT-22B and
Gemma~2~\citep{dehghani2023scaling,team2024gemma2}. RMSNorm provides an efficient
normalization form widely used in modern language models~\citep{zhang2019rmsnorm}.
QK-Clip, used in MuonClip, directly clips attention logits to improve stability at
large scale~\citep{kimik2}. We compare against QK-Clip because it targets the same
high-logit failure mode but acts after the dot product, whereas QK RMSNorm changes
the query/key geometry before logits are formed.

\paragraph{MLA with positional components.}
MLA separates a latent content path from a small materialized positional path, which
is typically implemented with rotary position embeddings
~\citep{su2021roformer,deepseekv2,deepseekv3}. Existing absorbed decode formulations
focus on avoiding full-key materialization by transforming queries into the latent key
space. Our contribution is to show that post-up-projection QK RMSNorm can be
expressed in the same absorbed form with only an additional scalar cache, while
keeping the RoPE component materialized and locally normalized.

\section{Limitations}

The empirical comparison uses 400M-parameter models trained for up to 100B
tokens---sufficient to test algebraic compatibility, implementation feasibility, and
behavior over a long token horizon, but not to establish frontier-scale behavior.
Larger models are needed to test whether the training-loss advantage persists at
billion-parameter scale. The logit-scale analysis is a mechanism-level explanation
rather than a complete optimization theory. Finally, although the transformation is
algebraically exact, practical speed depends on kernel integration: the per-query and
per-key scalar factors should be fused into the attention path to avoid unnecessary
memory traffic.

\section{Conclusion}

We showed that post-up-projection QK RMSNorm is compatible with efficient MLA
decoding. The static RMSNorm affine weights can be absorbed into the MLA query-side
projection, while the dynamic key RMS statistics become a small per-token,
per-KV-group scalar cache. This preserves MLA's latent cache while enabling QK
normalization as a training-stability mechanism. In our 400M runs trained for up to
100B tokens, \method{} achieves lower training loss and better downstream accuracy
than QK-Clip, and H800 decode benchmarks show less than $2\%$ latency overhead across
contexts up to 256k tokens.

\bibliography{iclr2026_conference}
\bibliographystyle{iclr2026_conference}

\appendix

\section{Experimental Configuration}
\label{app:experimental-config}

Table~\ref{tab:model-config} summarizes the model configuration used in our training
experiments. Table~\ref{tab:training-config} lists the few training and stabilization
hyperparameters that differ from the default recipe or are needed to reproduce the
comparison.

\begin{table}[h]
\caption{Model configuration.}
\label{tab:model-config}
\begin{center}
\small
\begin{tabular}{ccccc}
\toprule
Size & Layers & Heads & Hidden size & Head dim. $(d_c+d_r)$ \\
\midrule
400M & 12 & 16 & 1536 & $128+64$ \\
\bottomrule
\end{tabular}
\end{center}
\end{table}

\begin{table}[h]
\caption{Training and stabilization hyperparameters for the 400M runs.}
\label{tab:training-config}
\begin{center}
\small
\begin{tabular}{cc}
\toprule
Setting & Value \\
\midrule
Optimizer & Muon \\
Muon learning rate & $2\times 10^{-3}$ \\
LR schedule & WSD, minus-sqrt decay \\
QK-Clip threshold & 100 \\
QK-Clip $\alpha_{\mathrm{clip}}$ & 0.5 \\
\bottomrule
\end{tabular}
\end{center}
\end{table}

\section{Blockwise versus Unified QK RMSNorm}
\label{app:blockwise-vs-unified}

The main method uses blockwise QK RMSNorm because it matches the natural boundary
between MLA's latent content path and its small materialized RoPE path. A more literal
full-head variant would normalize the concatenated vector:
\[
    \hat q
    =
    \operatorname{RMSNorm}_q([q^{0,c};q^{0,r}]),
    \qquad
    \hat k
    =
    \operatorname{RMSNorm}_k([k^{0,c};k^{0,r}]).
\]
This remains algebraically compatible with MLA, since the key statistic is still one
scalar per token and KV group:
\[
    S^k_{t,g}
    =
    \sqrt{
        \frac{
            \lVert k^{0,c}_{t,g}\rVert_2^2
            +
            \lVert k^{0,r}_{t}\rVert_2^2
        }{
            d_c+d_r
        }
        + \epsilon
    }.
\]
However, unified normalization is less suitable as a training-time implementation
target in the efficient MLA stack considered here. We explain why below.

\paragraph{Training forward pass: synchronization point.}
In an efficient MLA training forward pass, the content path and the RoPE path are
computed by independent fused kernels with different tensor layouts, arithmetic
intensity, and memory-access patterns. The content path computes $c_t = x_t W_D$
and $K^{0,c}_t = c_t W^K$ via dense GEMMs. The RoPE path computes
$k^{0,r}_t = x_t W_{KR}$, applies rotary embeddings, and writes the small
materialized RoPE cache. These two paths proceed independently and can be overlapped
or fused into separate pipeline stages.

Unified key RMSNorm introduces a data dependency: the normalized content key
$\hat k^{0,c}_{t,g}$ cannot be written until the RoPE path produces
$\lVert k^{0,r}_t \rVert_2^2$. This forces a synchronization point between two
otherwise independent compute streams. In practice, this means either (1) serializing
the two paths, which eliminates kernel-fusion opportunities and exposes pipeline
bubbles, or (2) launching an additional synchronization kernel that waits for both
partial norms, which adds kernel-launch overhead and occupancy fragmentation on every
token at every layer. For a model with $L$ layers processing $T$ tokens, unified
normalization would insert $LT$ additional synchronization points into the forward
pass.

\paragraph{Score-computation coupling.}
Even setting aside cache-writer synchronization, unified normalization also makes score
computation less local. The final logit would take the form
\begin{equation}
    \ell_{i,t,h}
    =
    \left(
        \ell^{c,\mathrm{raw}}_{i,t,h}
        +
        \ell^{r,\mathrm{raw}}_{i,t,h}
    \right)
    \alpha^q_{i,h}
    \alpha^k_{t,g(h)},
    \label{eq:unified-logit}
\end{equation}
where $\ell^{c,\mathrm{raw}}$ is produced by an absorbed latent dot product and
$\ell^{r,\mathrm{raw}}$ by a materialized RoPE dot product. The scalar factor must be
applied after the content and RoPE reductions meet, rather than locally within each
component's natural score path. In fused attention kernels (e.g.\ FlashAttention-style
online softmax), this means the content and RoPE score tiles cannot be normalized
independently---they must be summed before the per-element scalar multiplication,
complicating the natural tiling strategy.

\paragraph{Summary.}
Blockwise normalization avoids both costs: the content RMS statistic depends only
on $c_t W^K_g$ (available within the content GEMM path), and the RoPE vector is
materialized and can be normalized directly with no cross-path dependency. This is an
implementation constraint in our target stack: unified QK RMSNorm does not fit as
cleanly with the fused, pipelined kernel structure that makes MLA training efficient.
We therefore use blockwise QK RMSNorm as the implementation-compatible choice rather
than treating unified normalization as the primary ablation target.

\section{Optimization Diagnostics (400M / 100B)}
\label{app:diagnostics-100b}

\begin{figure}[h]
\begin{center}
\includegraphics[width=0.6\linewidth]{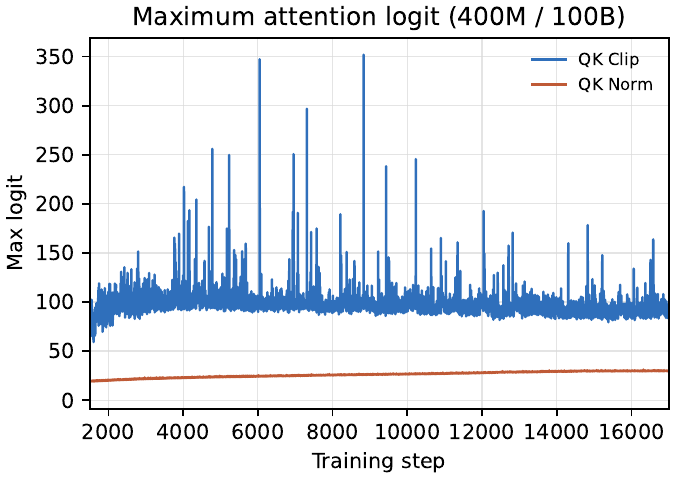}
\end{center}
\caption{Maximum attention logit for the 400M/100B run.
QK-Clip fluctuates near its clipping threshold; \method{} remains lower.}
\label{fig:max-logit-100b}
\end{figure}

\begin{figure}[h]
\begin{center}
\includegraphics[width=0.6\linewidth]{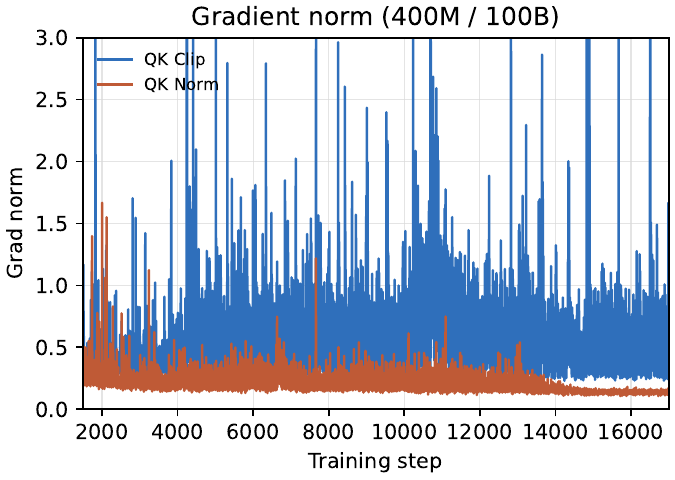}
\end{center}
\caption{Gradient norm for the 400M/100B run.
\method{} exhibits fewer large spikes than QK-Clip. The y-axis is capped at 3 for
readability.}
\label{fig:grad-norm-100b}
\end{figure}

Figures~\ref{fig:max-logit-100b} and~\ref{fig:grad-norm-100b} show the maximum
attention logit and gradient norm for the 400M/100B runs. \method{} keeps logits in a
lower range than QK-Clip and produces a smoother gradient-norm profile,
consistent with the mechanism described in Section~4.2.

\section{QK-Normed Max-Logit Growth}
\label{app:qknorm-max-logit}

\begin{figure}[h]
\begin{center}
\includegraphics[width=0.6\linewidth]{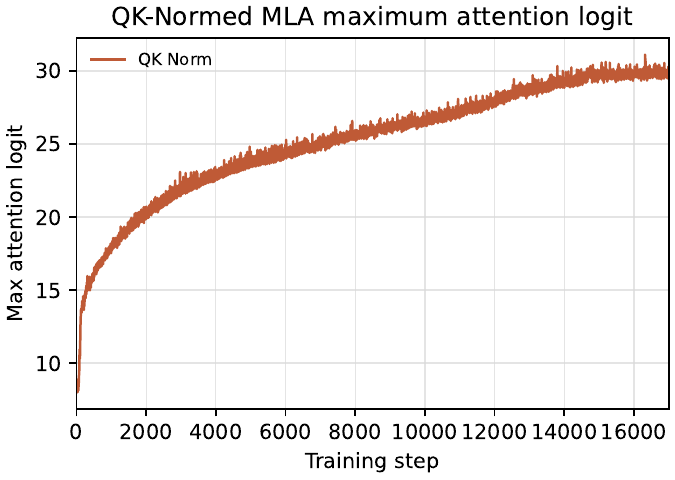}
\end{center}
\caption{Maximum attention logit of \method{} over training. The growth is sublinear
and converges, consistent with affine-weight growth rather than raw query/key norm
growth.}
\label{fig:qknorm-max-logit}
\end{figure}

Figure~\ref{fig:qknorm-max-logit} shows the maximum attention logit of \method{}
over training. The curve grows sublinearly in early training and converges to a stable
plateau, consistent with QK normalization controlling attention-logit scale.
The residual growth is expected: RMSNorm removes the dependence on raw query and key
RMS values, but the learned affine weights $\gamma_q\odot\gamma_k$ can still grow
slowly during training.

For one block of QK RMSNorm, write
\[
    \hat q
    =
    \frac{q^0\odot\gamma_q}{S(q^0)},
    \qquad
    \hat k
    =
    \frac{k^0\odot\gamma_k}{S(k^0)}.
\]
Let
\[
    u_q = \frac{q^0}{S(q^0)}, \qquad
    u_k = \frac{k^0}{S(k^0)}.
\]
The normalized logit can be written as
\[
    \ell
    =
    \hat q^\top \hat k
    =
    \sum_{j=1}^{d}
    u_{q,j}u_{k,j}\gamma_{q,j}\gamma_{k,j}.
\]
Therefore
\[
    |\ell|
    \le
    \sum_{j=1}^{d}
    |u_{q,j}u_{k,j}|
    |\gamma_{q,j}\gamma_{k,j}|.
\]
Since the RMS-normalized directions $u_q$ and $u_k$ have controlled RMS, the remaining
learned source of logit-scale growth is the affine product
$\gamma_q\odot\gamma_k$. Thus a slow increase in the maximum logit is expected as
parameter norms and RMSNorm weights grow during training. The important distinction
is that the growth is mediated by learned affine weights rather than by uncontrolled
raw query/key norms.

\section{Reference Equivalence Test}
\label{app:equivalence-test}

For implementation testing, we compare two paths on random tensors.
The explicit path computes
\[
    k^{0,c}_{t,g}=c_tW^K_g,
    \qquad
    \hat q^c_{i,h}
    =
    \frac{q^{0,c}_{i,h}\odot\gamma^c_q}{S^q_{i,h}},
    \qquad
    \hat k^c_{t,g}
    =
    \frac{k^{0,c}_{t,g}\odot\gamma^c_k}{S^k_{t,g}},
\]
and then forms
\[
    \ell^c_{i,t,h}
    =
    \hat q^c_{i,h}\cdot\hat k^c_{t,g(h)}.
\]
The absorbed path computes
\[
    \tilde q^c_{i,h}
    =
    (q^{0,c}_{i,h}\odot\gamma^c_q\odot\gamma^c_k)
    (W^K_{g(h)})^\top,
\]
and then
\[
    \ell^c_{i,t,h}
    =
    (\tilde q^c_{i,h}\cdot c_t)
    \alpha^q_{i,h}
    \alpha^k_{t,g(h)}.
\]
The maximum absolute difference between the two logits should be at floating-point
roundoff level.

\end{document}